\documentclass{article}

\usepackage{microtype}
\usepackage{graphicx}
\usepackage{subfigure}
\usepackage{booktabs} 

\usepackage{hyperref}

\usepackage[accepted]{icml2021}

\usepackage{times}
\usepackage{xcolor}
\usepackage{soul}
\usepackage[utf8]{inputenc}
\usepackage[T1]{fontenc}    
\usepackage{color}
\usepackage{epsfig}
\usepackage{amsthm}
\usepackage{amsmath}
\usepackage{amssymb}
\usepackage{float}
\usepackage{graphicx}
\usepackage{wrapfig}
\usepackage{makecell}
\usepackage{caption}
\usepackage{bm}
\usepackage{multirow}
\usepackage{microtype}
\usepackage{mathtools}
\usepackage{enumitem}
\usepackage{ulem}
\usepackage{xcolor}
\usepackage{tabularx}

\newcommand{\ie}[1]{{\textit{i.e.}}}
\newcommand{\eg}[1]{{\textit{e.g.}}}

\usepackage{commands}

\icmltitlerunning{Defending against Reconstruction Attack  in Vertical Federated Learning}

\begin{document}

\twocolumn[
\icmltitle{Defending against Reconstruction Attack  in Vertical Federated Learning}



\icmlsetsymbol{equal}{*}

\begin{icmlauthorlist}
\icmlauthor{Jiankai Sun}{us-sea}
\icmlauthor{Yuanshun Yao}{us-ca}
\icmlauthor{Weihao Gao}{cn}
\icmlauthor{Junyuan Xie}{cn}
\icmlauthor{Chong Wang}{us-sea}
\end{icmlauthorlist}

\icmlaffiliation{us-sea}{Bytedance Inc., Seattle, USA.}
\icmlaffiliation{us-ca}{Bytedance Inc., Mountain View, USA.}

\icmlaffiliation{cn}{Bytedance Ltd., Beijing, China.}

\icmlcorrespondingauthor{Jiankai Sun}{jiankai.sun@bytedance.com}
\icmlcorrespondingauthor{Chong Wang}{chong.wang@bytedance.com}

\icmlkeywords{Machine Learning, ICML}

\vskip 0.3in
]



\printAffiliationsAndNotice{} 

\begin{abstract}
Recently researchers have studied input leakage problems in \textit{Federated Learning} (FL) where a malicious party can reconstruct sensitive training inputs provided by users from shared gradient~\cite{zhu2019dlg,geiping2020inverting,yin2021see}. It raises concerns about FL since input leakage contradicts the privacy-preserving intention of using FL. Despite relatively rich literature on attacks and defenses of input reconstruction in Horizontal FL, input leakage and protection in vertical FL starts to draw researchers attention recently. In this paper, we study how to defend against input leakage attack in Vertical FL. We design an adversarial training based framework that contains three modules: adversarial reconstruction, noise regularization, and distance correlation minimization. Those modules can not only be employed individually but also applied together since they are independent to each other. Through extensive experiments on a large-scale industrial online advertising dataset, we show our framework is effective in protecting input privacy while retaining the model utility. 

\end{abstract}

\section{Introduction}

With the increasing concerns on data security and user privacy in machine learning, \textit{Federated Learning} (FL)~\citep{mcmahan2017communication} becomes a promising solution to allow multiple parties collaborate without sharing their data completely. Based on how sensitive data are distributed among various parties, FL can be classified into two categories~\citep{yang2019federated}: Cross-silo or Vertical FL (\textit{vFL}) and Cross-device or Horizontal FL (\textit{hFL}). 
In contrast to hFL where the data are partitioned by samples or entities (i.e. a person), vFL partitions the data by different attributes (i.e. features and labels). In vFL, multiple parties can own different attributes from the same entities.

One typical example of vFL is a collaboration between general and specialized hospitals. They might hold the data for the same patient, but the general hospital owns generic information (\textit{i.e.} private attributes such as gender and age) of the patient while the specialized hospital owns the specific testing results (\textit{i.e.} labels) of the same patient. Therefore they can use vFL to jointly train a model that predicts a specific disease examined by the specialized hospital from the features provided by the general hospital.

Under two-party vFL setting, the model is split into two submodels and each submodel is owned by one party. During training, the party without labels (namely \textit{passive party}) sends the computation results (namely \textit{embedding}) of the intermediate layer (namely \textit{cut layer}) rather than the raw data to the party with labels (namely \textit{active party}). The active party takes the embedding as the input, completes the rest of forward pass, computes backward gradient based on the labels, and performs backward pass up to the cut layer. Then it sends the gradient w.r.t the cut layer back to the passive party. Finally the passive party completes the backpropagation with the gradients of the cut layer using chain rule. 

At first glance, vFL seems private because no feature/input or label is shared between the two parties. 
However, from the viewpoint of passive party, the cut layer embedding still contains rich information which can be exploited by a malicious active party to leak the input information. Recently researchers have identified some input leakage problems under hFL settings. For example, Mahendran et al. showed that an attacker can exploit the intermediate embedding to reconstruct the input images, and hence the people who show up in the input images can be re-identified~\cite{understandingDeepImage2015}. 
Furthermore, ~\citeauthor{zhu2019dlg,geiping2020inverting,yin2021see} showed that in hFL setting, the central server could recover the raw inputs and labels of the clients from the gradient sent from clients.


Despite the relatively well-studied problems of input leakage in hFL, defending against reconstruction attack starts to draw researchers attention recently ~\cite{featureinference2021}. 
In this paper, we propose an adversarial training based framework that can defend against input reconstruction attack in vFL. The proposed framework simulates the game between an attacker (\ie{} the active party) who actively reconstructs raw input from the cut layer embedding and a defender (\ie{} the passive party) who aims to prevent the input leakage. Our framework consists of three modules to protect input privacy: adversarial reconstructor, noise regularization, and distance correlation. These modules are designed to make the submodel owned by the passive party more robust against potential attacks that extract sensitive information about the raw input from
the cut layer embedding. The adversarial reconstructor is designed to maximize the reconstruction error of the attacker. Noise regularization is designed to reduce information about input in embedding by misleading attacker's optimization toward a random direction. Distance correlation module is to decrease the correlation between the raw input and the cut layer embedding. We conduct extensive experiments in a large-scale online advertising dataset collected under an industrial setting to demonstrate the effectiveness of our framework in protecting input privacy while retaining model utility.

We summarize our contributions as follows:

\begin{itemize}
    \item We design an adversarial training based framework with three independent modules to defend against input reconstruction attack in vFL.
    \item Through extensive experiments on a real-world and industrial-scale online advertising dataset, we show our framework can achieve a good trade-off between preserving input privacy and retaining model performance.
    
\end{itemize}
\section{Methodology}
\label{sec:framework}



\begin{figure*}[h!]
    \includegraphics[width=\textwidth]{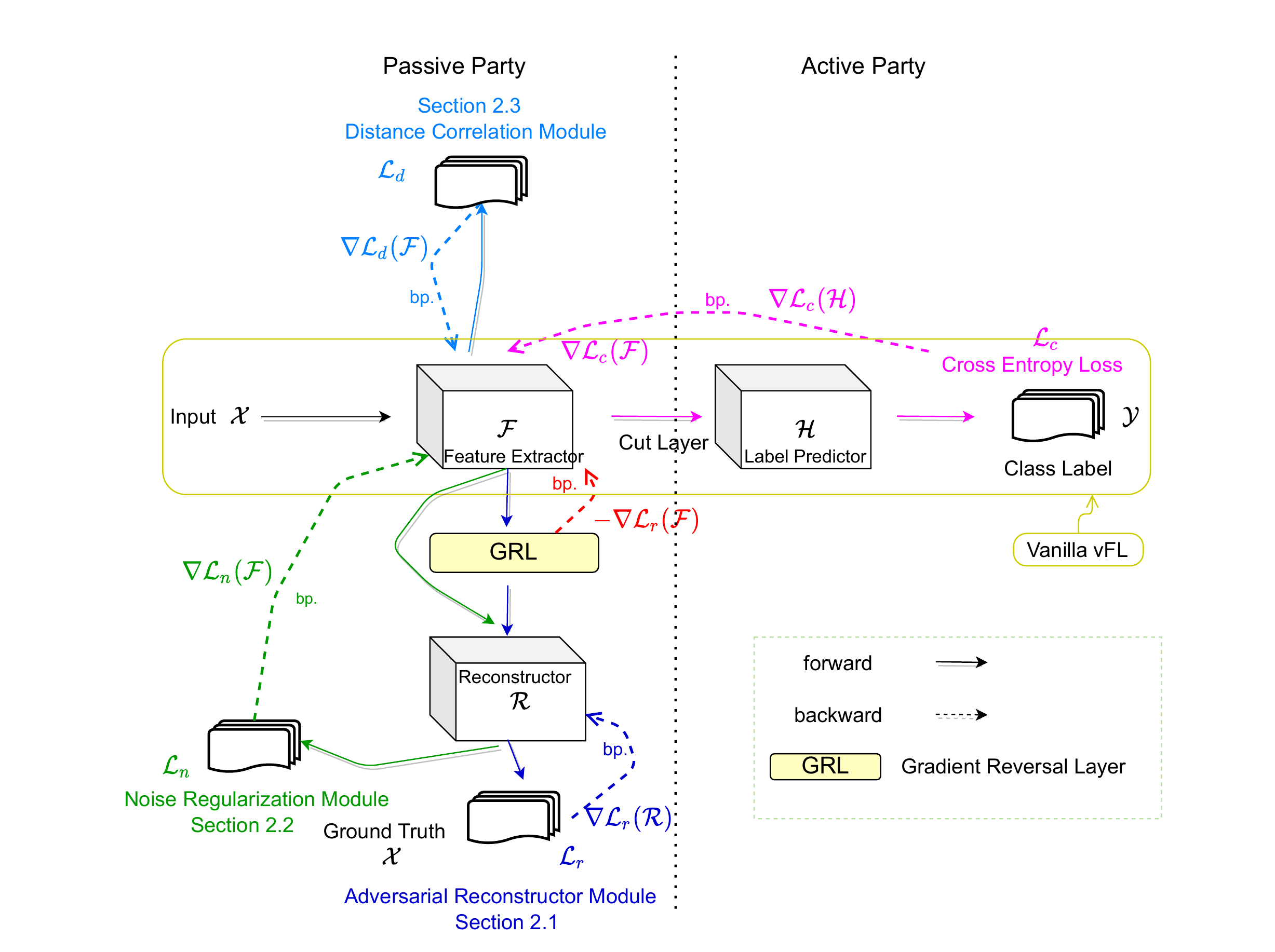}
    \caption{Overview of our framework DRAVL. Vanilla vFL consists of two classical modules: feature extractor $\mathcal{F}$ and label predictor $\mathcal{H}$. DRAVL contains three additional privacy related modules in the passive party side: Adversarial Module (~\ref{subsec:ar}), Noise Regularization Module (Section ~\ref{subsec:nr}), and Distance Correlation Module (Section ~\ref{subsec:dcor}).  
    }
    \label{fig:framework_overview}
\end{figure*}

\paragraph{vFL Background.} We begin by providing some background on how the vanilla vFL works, as shown in the part of Figure~\ref{fig:framework_overview}. A conventional vFL framework splits the model into two parts: feature extractor $\mathcal{F(.)}$ (owned by passive party) and label predictor $\mathcal{H(.)}$ (owned by active party). In the forward pass, the passive party feeds the raw input $\mathcal{X}$ into $\mathcal{F(.)}$, and then sends the cut layer embedding $\mathcal{F(X)}$ to the active party. The active party takes $\mathcal{F(X)}$ as the input for the label predictor $\mathcal{H(.)}$ (designed for the intended task), and then computes the gradients based on the ground-truth labels $\mathcal{Y}$ owned by it. Then, in the backward pass, the active party sends the gradient with respect to the cut layer back ($\frac{\partial{\mathcal{L}}}{\partial{\mathcal{F}}}$) to the passive party. Finally, the passive party completes the backpropagation using chain rule and updates $\mathcal{F(.)}$. 


\paragraph{Threat Model.} We assume the attacker is a malicious active party that attempts to reconstruct input $\mathcal{X}$ from the cut layer embedding $\mathcal{F(X)}$ passed by the passive party. The attacker has access to $\mathcal{F(X)}$ and label $\mathcal{Y}$. Our goal, as the defender and the passive party, is to prevent the reconstruction by making feature extractor $\mathcal{F(.)}$ more robust. We have access to $\mathcal{F(X)}$, raw input $\mathcal{X}$, and ability to modify feature extractor $\mathcal{F(.)}$.
 
\paragraph{Framework Overview.} Figure~\ref{fig:framework_overview} shows the design of our framework which consists of three modules: Adversary Reconstructor (\texttt{AR}), Noise Regularization (\texttt{NR}), and Distance Correlation (\texttt{dCor}). These modules are designed to hide privacy-sensitive information from $\mathcal{F(.)}$ that can be exploited by the attacker (\ie{} active party) to reconstruct the raw input $\mathcal{X}$ from $\mathcal{F(X)}$. Specifically, AR (Section~\ref{subsec:ar}) is designed to simulate an attacker that actively attempts to reconstruct the input, and then it maximizes error of the attacker. NR (Section~\ref{subsec:nr}) is designed to reduce information about $\mathcal{X}$ in $\mathcal{F(.)}$ and stabilize AR. dCor (Section~\ref{subsec:dcor}) is designed to decrease the correlation between $\mathcal{X}$ and $\mathcal{F(X)}$. Note that since these three modules are independent to each other, they can be either implemented as separate modules or unified into a single framework. We name this united framework as \texttt{DRAVL} (\textbf{D}efending against \textbf{R}econstruction \textbf{A}ttack in \textbf{V}ertical Federated \textbf{L}earning).


\subsection{Adversarial Reconstructor Module}
\label{subsec:ar}

Inspired by prior work~\cite{LiDeepObfuscator,feutry2018learning,GAN2014}, AR simulates an adversarial attacker who aims to train a reconstructor $\mathcal{R(.)}$ that maps the embedding $\mathcal{F(X)}$ to the input $\mathcal{X}$ by minimizing the following reconstruction loss:

\begin{equation}
\label{eq:ar}
\mathcal{L}_r = ||\mathcal{R(F(X)) - \mathcal{X}})||_2^2
\end{equation}

where $\mathcal{R(.)}$ can be any model (\eg{} a MLP).

Ideally we can formulate the protection as a max-min problem that maximizes the minimized $\mathcal{L}_r$. However we empirically find that such optimization is unstable and hard to tune. Instead, we use \textit{Gradient Reversal Layer} (GRL) from prior work~\cite{grl2015,feutry2018learning} that demonstrated promising results in stabilizing adversarial training. As shown in Figure~\ref{fig:framework_overview}, GRL is inserted between the feature extractor $\mathcal{F(.)}$ and the adversarial reconstructor $\mathcal{R(.)}$. In forward pass, GRL just performs identity transformation. In backward pass, it multiplies the corresponding gradient w.r.t to the cut layer by $-\lambda$ ($\lambda > 0$, i.e. $\lambda = 1$) and passes $- \lambda \frac{\partial{\mathcal{L}_r}}{\partial{\mathcal{F}}}$ to the preceding layer. Intuitively, GRL leads to the opposite of gradient descent that is performing gradient ascent on the feature extractor $\mathcal{F(.)}$ with respect to maximize the adversarial reconstruction loss (as an attacker). Therefore it roughly achieves the goal of maximizing the minimized reconstruction loss. After inserting GRL, we just need to minimize $\mathcal{L}_r$ as adversarial training. We update both $\mathcal{F(.)}$ and $\mathcal{R(.)}$ during training; after training is finished, we discard $\mathcal{R(.)}$ and save $\mathcal{F(.)}$ as the more robust feature extractor.

\subsection{Noise Regularization Module}
\label{subsec:nr}



Noise Regularization also simulates an adversarial reconstructor $\mathcal{R}'(.)$ but with a different goal. It is designed to reduce information about $\mathcal{X}$ in $\mathcal{F(X)}$ by misleading the reconstructed input $\mathcal{R'(F(X))}$ toward a random direction (and therefore degrading reconstruction quality). During training, we generate random Gaussian noise $\mathcal{N}_{noise}$\footnote{Random noise from other distributions such as Uniform distribution are effective too.} and minimize the following noise regularization loss:

\begin{equation}
\mathcal{L}_n = ||\mathcal{R'(F(X))} - \mathcal{N}_{noise}||_2^2
\end{equation}

Note that if only NR is used, we can train an independent reconstructor (without GRL) as $\mathcal{R(.)}'$. If AR and NR are used together, we can reuse the AR reconstructor $\mathcal{R(.)}$ as $\mathcal{R(.)}'$ \footnote{Since the adversarial training is effective, $\mathcal{R(.)}$ and $\mathcal{R(.)}'$ can achieve similar reconstruction performance.} 
. In our experiments, we use the same reconstructor for both AR and NR. For notation simplicity, we will use $\mathcal{R(.)}$ to represent both AR and NR reconstructor.

As shown in Figure~\ref{fig:framework_overview}, $\mathcal{F(X)}$ are fed into the reconstructor $R$ directly (without GRL). NR calculates $\mathcal{L}_n$ and computes the gradients of the $\mathcal{L}_n$ w.r.t. $\mathcal{F(.)}$, \ie{} $\nabla \mathcal{L}_n(\mathcal{F})$, and updates  $\mathcal{F(.)}$ accordingly via backpropogation. 
Note that $\mathcal{R(.)}$ only works as a reconstructor and provides the reconstructed result $\mathcal{R(F(X))}$ as an input for $\mathcal{L}_n$ optimized by NR. NR only has effects on $\mathcal{F(.)}$ and does not update any parameters of $\mathcal{R}$ during backpropogation phase.  


\subsection{Distance Correlation Module}
\label{subsec:dcor}

Distance correlation is designed to make input $\mathcal{X}$ and embedding $\mathcal{F(X)}$ less dependent, and therefore reduces the likelihood of gaining information of $\mathcal{X}$ from $\mathcal{F(X)}$. Distance correlation  measures statistical dependence between two vectors\footnote{Two vectors can have different length. }~\cite{nopeek2018}. The distance correlation loss is the following:

\begin{equation}
\mathcal{L}_d = \log{(dCor(\mathcal{X}, \mathcal{F(X)))}}
\end{equation}

where we minimize the (log of) distance correlation loss ($\mathcal{L}_d$) during the model training. It can be interpreted as $\mathcal{X}$ being a good proxy dataset to construct $\mathcal{F(X)}$ but not as vice versa in terms of reconstructing $\mathcal{X}$ from $\mathcal{F(X)}$ ~\cite{nopeek2018}.


Note that dCor computates pairwise distance between samples and requires $O(n^2)$ time complexity where $n$ is the batch size. In practice, there are some faster estimators of dCor~\cite{fastdcor2019,huang2017statistically}. In addition, dCor is sensitive to the $n$ and a larger $n$ can give a more accurate estimation of the distance correlation. 



\subsection{A Unified Framework}
We can unify all three modules into one framework. The overall loss function is a combination of four losses: adversarial reconstruction loss ($\mathcal{L}_r$), noise regularization loss ($\mathcal{L}_n$), distance correlation loss ($\mathcal{L}_d$), and normal label prediction loss  ($\mathcal{L}_c$). In this paper, we focus on classification and use categorical cross entropy as label prediction loss. Optimizing $\mathcal{L}_c$ makes sure the model maintain a good utility; optimizing $\mathcal{L}_r$, $\mathcal{L}_n$, and $\mathcal{L}_d$ increases model privacy. Uniting them in one framework can help us defend against the reconstruction attack while maintaining the accuracy of the primary learning task. The overall loss function is:


\begin{equation}
\mathcal{L} = \mathcal{L}_c + \alpha_r \mathcal{L}_r + \alpha_n \mathcal{L}_n  + \alpha_d \mathcal{L}_d 
\end{equation}

where $\alpha_d \geq 0$, $\alpha_n \geq 0$ and $\alpha_r \geq 0$ are weights for distance correlation, noise regularization, and adversarial reconstructor module respectively. Note that during training, only the passive party optimizes these three modules while there is no change on the active party's training optimization.
\section{Experimental Study}
\label{sec:experiment}

\paragraph{Dataset and Setting.} We evaluate the proposed framework on a large-scale industrial binary classification dataset for conversion prediction tasks  with millions of user click records. The data was collected over a period of three months from one of the largest online media platforms in industry (with hundreds of millions of users) that collaborates with e-commerce advertising. In total, the dataset contains $>42.56$ million records of user conversion interactions (samples). 


In our setting, the passive party is an online media platform that displays advertisements for an e-commerce company (the active party) to its users. 
Both parties have different attributes for the same set of users: the passive party has features of user viewing history on the platform and the active party has features of user product browsing history on its website and labels indicating if the user converted or not. Under our threat model, the goal of the passive party is to prevent their raw input features from being reconstructed from cut layer embedding.


\paragraph{Model.} We train a Wide\&Deep model ~\cite{cheng2016wide} where the passive party's feature extractor $\mathcal{F(.)}$ consists of the embedding layers for the input features and several layers of ReLU activated MLP (deep part) and the active party's label predictor $\mathcal{H(.)}$  consists of the last logit layer of the deep part and the entire wide part of the model. During training, in each batch the passive party sends an embedding matrix with size $512\times64$ to the passive party where batch size is $512$\footnote{The batch size is 512 if not specified.} and embedding size is $64$.

\paragraph{Evaluation Metrics.} To measure the privacy, we train an independent reconstructor $\mathcal{R(.)}_I$ that minimizes eq~\ref{eq:ar}\footnote{In practice, the attacker cannot train $\mathcal{R(.)}_I$ since as the passive party, he does not have access to the ground truth input $\mathcal{X}$. For evaluation purpose, our experiment simulates the most powerful attacker, and therefore the privacy-preserving performance when facing real attacks should only be higher than what we report.}. Note that $\mathcal{R(.)}_I$ is different from $\mathcal{R(.)}$ in AR or $\mathcal{R'(.)}$ in NR; it is the simulated attack used for evaluation purpose and agnostic during our defense. 
The input privacy is measured by $\mathcal{R(.)}_I$'s reconstruction loss, \ie{} the mean squared error (MSE) between its reconstructed input and real input\footnote{Different from existing reconstruction attacks on image models, our input is a set of user related features rather than humanly perceptible images. Therefore it is infeasible to show how well the reconstructed inputs look like visually. Instead we use MSE to quantify the reconstruction quality.}. A larger MSE means more privacy is preserved. We also measure the privacy with $dCor(\mathcal{X}, \mathcal{F(X)})$ as described in Section~\ref{subsec:dcor}. A lower dCor means less dependency between $\mathcal{X}$ and $\mathcal{F(X)}$ and therefore better privacy. To measure the model utility, we use AUC of conversion prediction. We use the online stream training to train the vFL model along with our framework. We average dCor, MSE and AUC in a daily basis and report them on evaluation data from Jan to Feb 2020. 


We first show experimental results of optimizing individual NR and dCor module alone. Then we combine them with AR as a united framework (DRAVL) and measure its performance.




\begin{figure*}[ht!]
  \begin{minipage}[b]{0.245\linewidth}
  \centering
    \includegraphics[width=\linewidth]{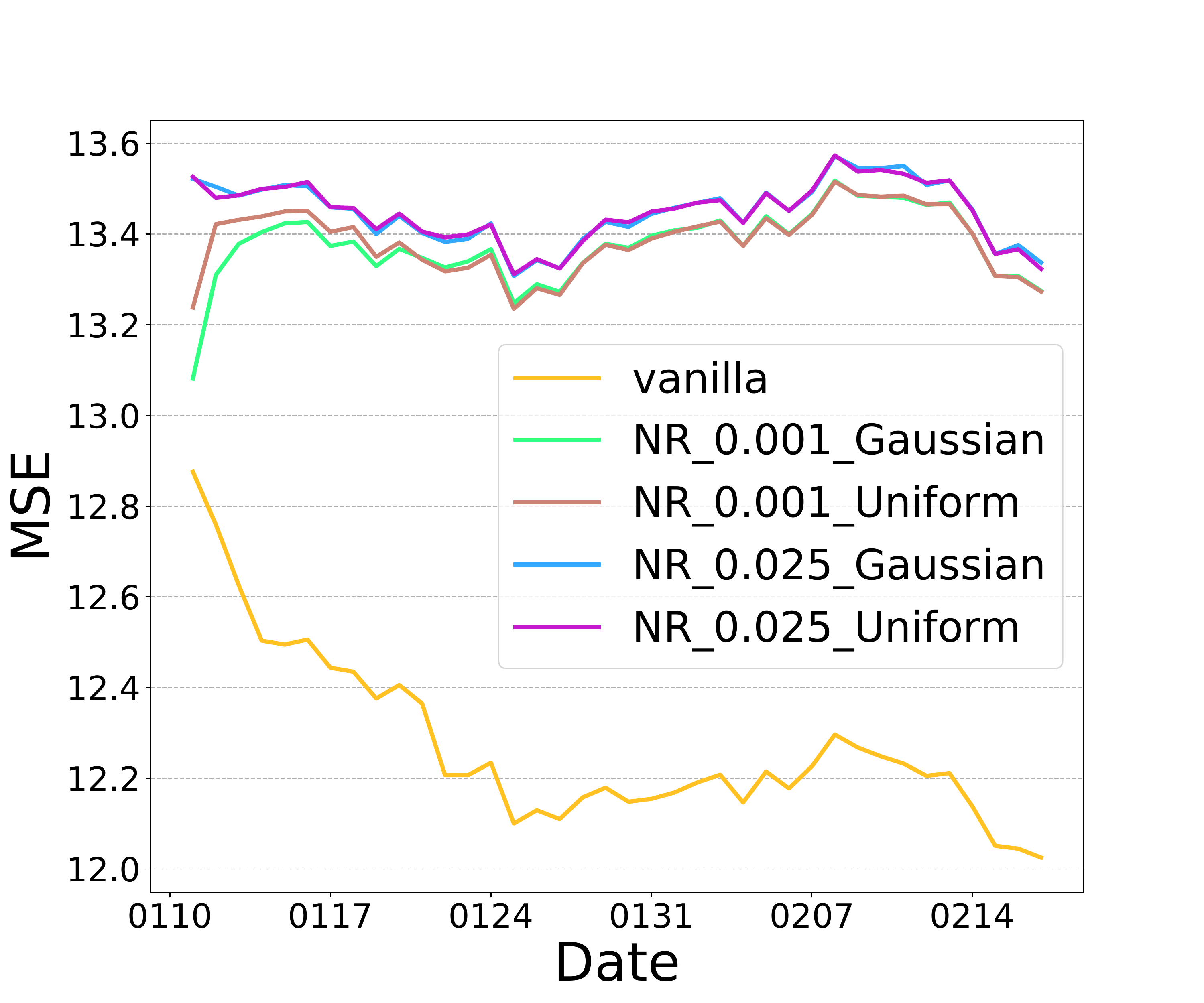}
    \caption*{(a: MSE of NR Module)}
  \end{minipage}
  \begin{minipage}[b]{0.245\linewidth}
  \centering
    \includegraphics[width=\linewidth]{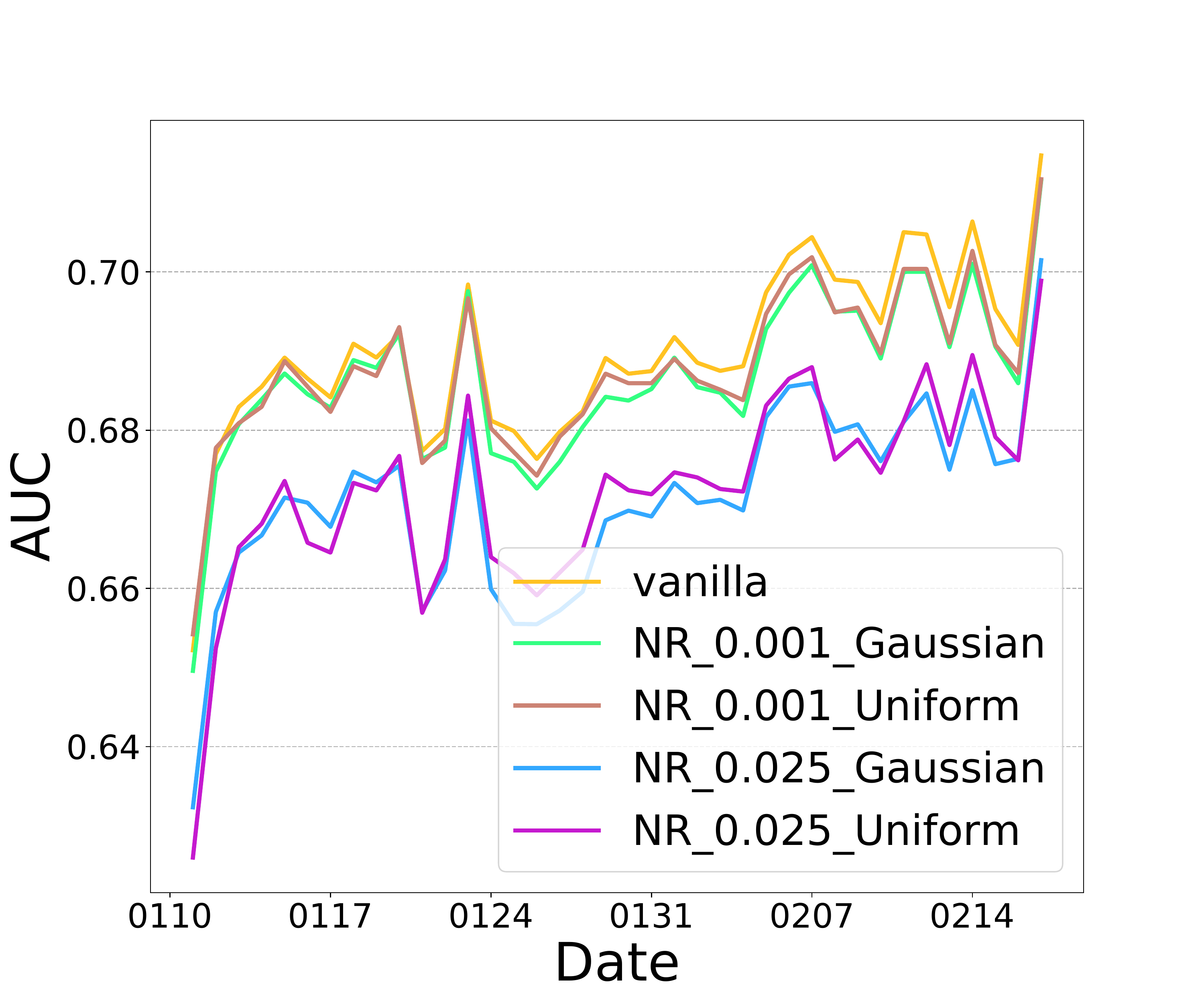}
    \caption*{(b: AUC of NR Module)}
  \end{minipage}
   \begin{minipage}[b]{0.245\linewidth}
  \centering
  \centering
    \includegraphics[width=\linewidth]{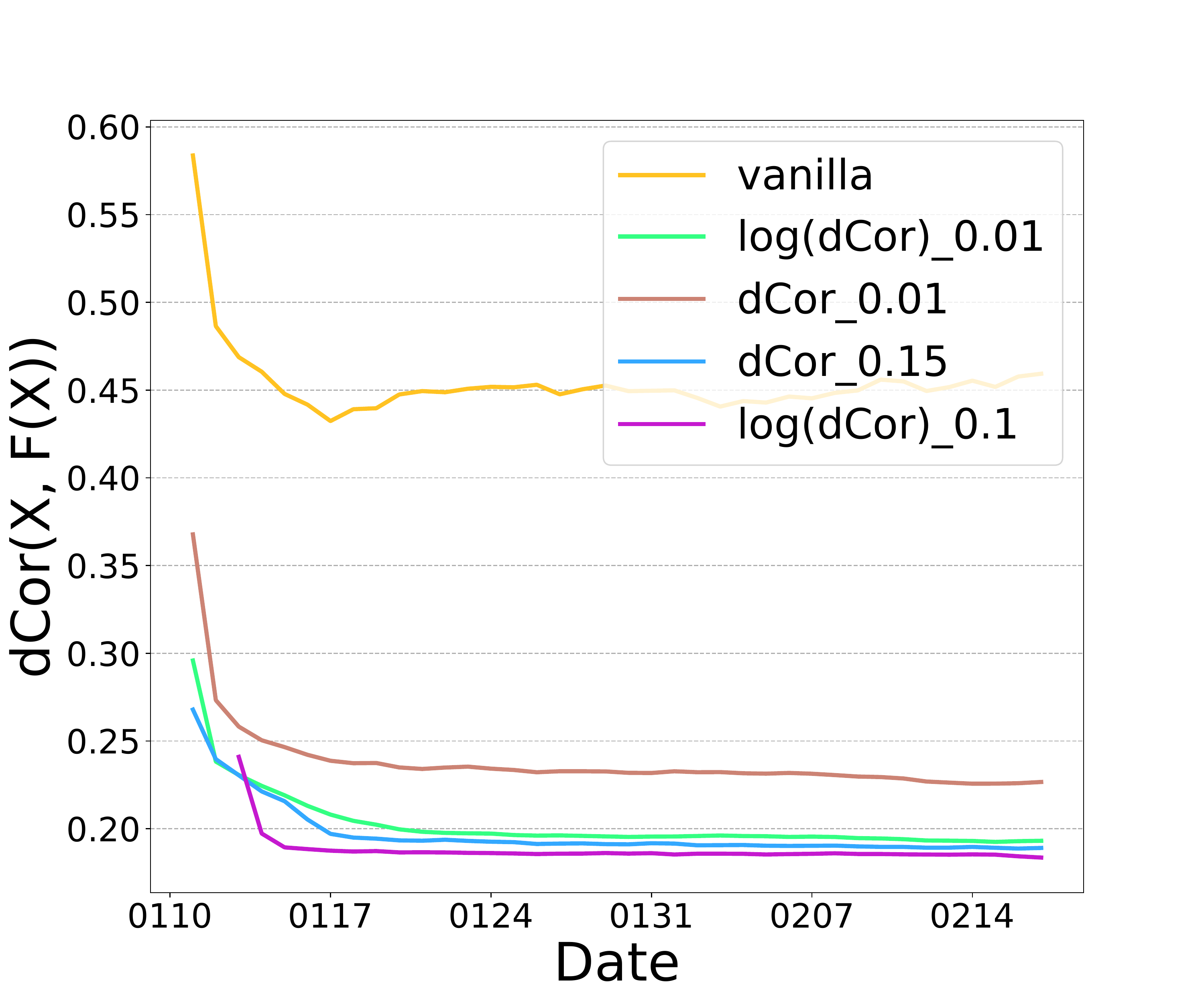}
    \caption*{(c: dCor of dCor Module)}
  \end{minipage}
  \begin{minipage}[b]{0.245\linewidth}
  \centering
    \includegraphics[width=\linewidth]{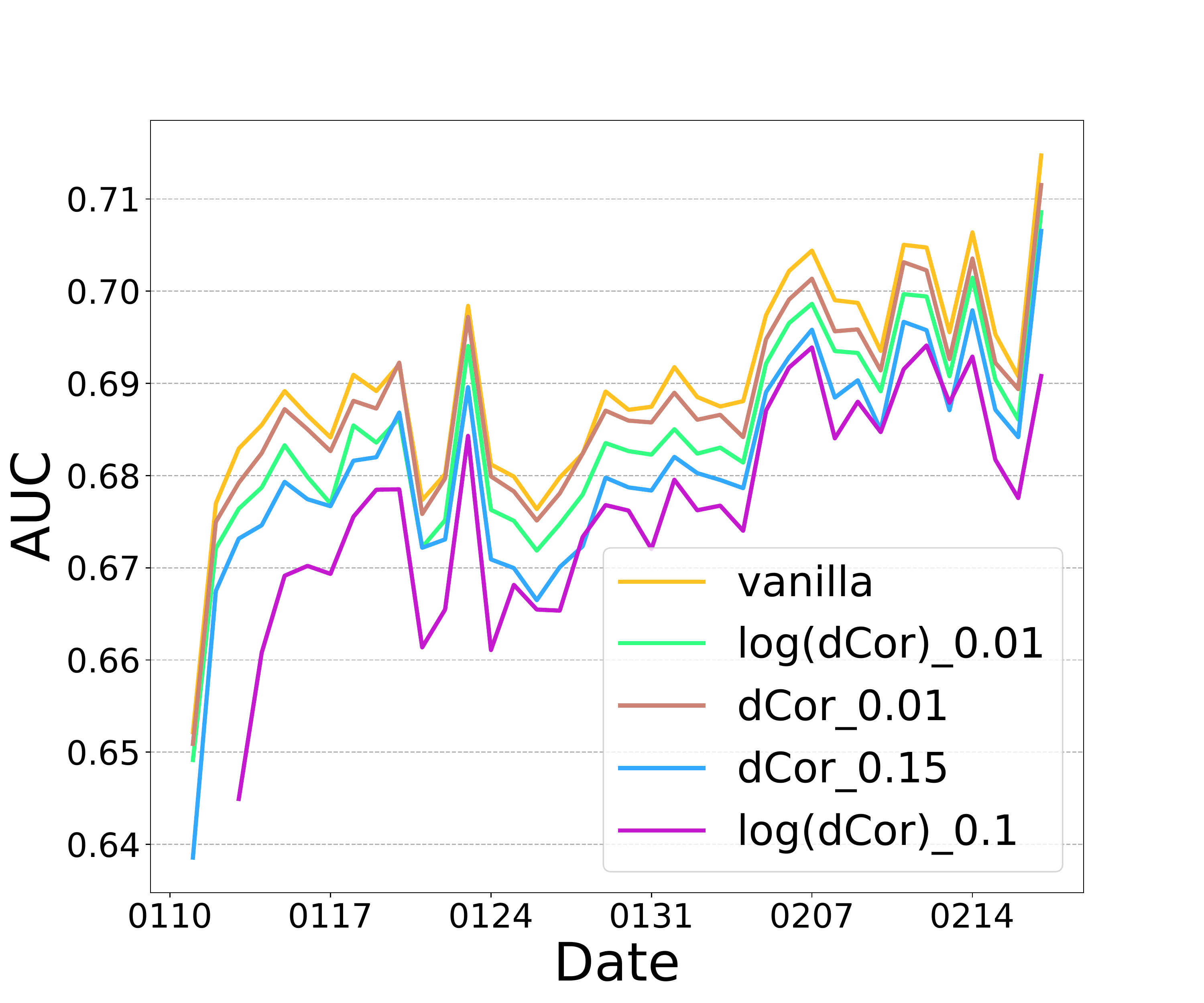}
    \caption*{(d: AUC of dCor Module)}
  \end{minipage}
 \caption{
  Figure (a) and (b) show MSE and AUC of optimizing NR module with different values of $\alpha_n$. Figure (c) and (d) show the results of optimizing dCor module with different values of $\alpha_d$. Figure (c) shows $dCor(\mathcal{X}, \mathcal{F(X)})$. Figure (d) shows the model AUC. $log(dCor)$ represents that we use $log(dCor(\mathcal{X}, \mathcal{F(X)}))$ in the loss function. 
 }
 \label{fig:minimizing_dcor_nr_individually} 
 \vspace{-0.15in}
\end{figure*}


\subsection{Evaluating Noise Regularization Module}
We evaluate the performance of using noise regularization module alone, \ie{} minimizing $\mathcal{L}_c + \alpha_n \mathcal{L}_n$. First, we evaluate the impact of noise choice on model privacy and utility. In Figure~\ref{fig:minimizing_dcor_nr_individually} (a) and (b), we compare two different types of random noise: Gaussian noise 
and Uniform noise 
. With the same $\alpha_r$, both random noises achieve similar MSE and AUC. Therefore the NR module is not sensitive to the type of the random noise. In addition, we can see the tradeoff between privacy (MSE) and utility (AUC) by varying the value of $\alpha_r$. A larger $\alpha_r$ leads to a higher MSE (therefore better privacy) but a lower AUC score (therefore worse utility). 


\subsection{Evaluating Distance Correlation Module}

We evaluate the performance of minimizing distance correlation alone, \ie{} minimizing $\mathcal{L} = \mathcal{L}_c + \alpha_d \mathcal{L}_d$. Figure~\ref{fig:minimizing_dcor_nr_individually} (c) and (d) show the dCor performance with different values of $\alpha_d$. First, \texttt{vanilla} vFL  without any privacy protection, can naturally reduce the $dCor(\mathcal{X}, \mathcal{F(X)})$ during the training (dropping from $0.6$ at the beginning of the training to $0.45$). Second, unsurprisingly optimizing dCor can reduce $dCor(\mathcal{X}, \mathcal{F(X)})$ more than vanilla ($0.45$ for vanilla and $0.2$ for dCor with $\alpha_d = 0.1$) while AUC drops less than $0.01$. It indicates that dCor module can achieve a reasonable privacy-utility tradeoff with an appropriate $\alpha_d$. Third,  log(dCor) is more robust to $\alpha_d$
than dCor since the gap of log(dCor) between $\alpha_d = 0.01$ and $\alpha_d = 0.1$ is much smaller.



\subsection{Effectiveness of DRAVL}

\begin{figure*}[ht!]
  \begin{minipage}[b]{0.33\linewidth}
  \centering
    \includegraphics[width=\linewidth]{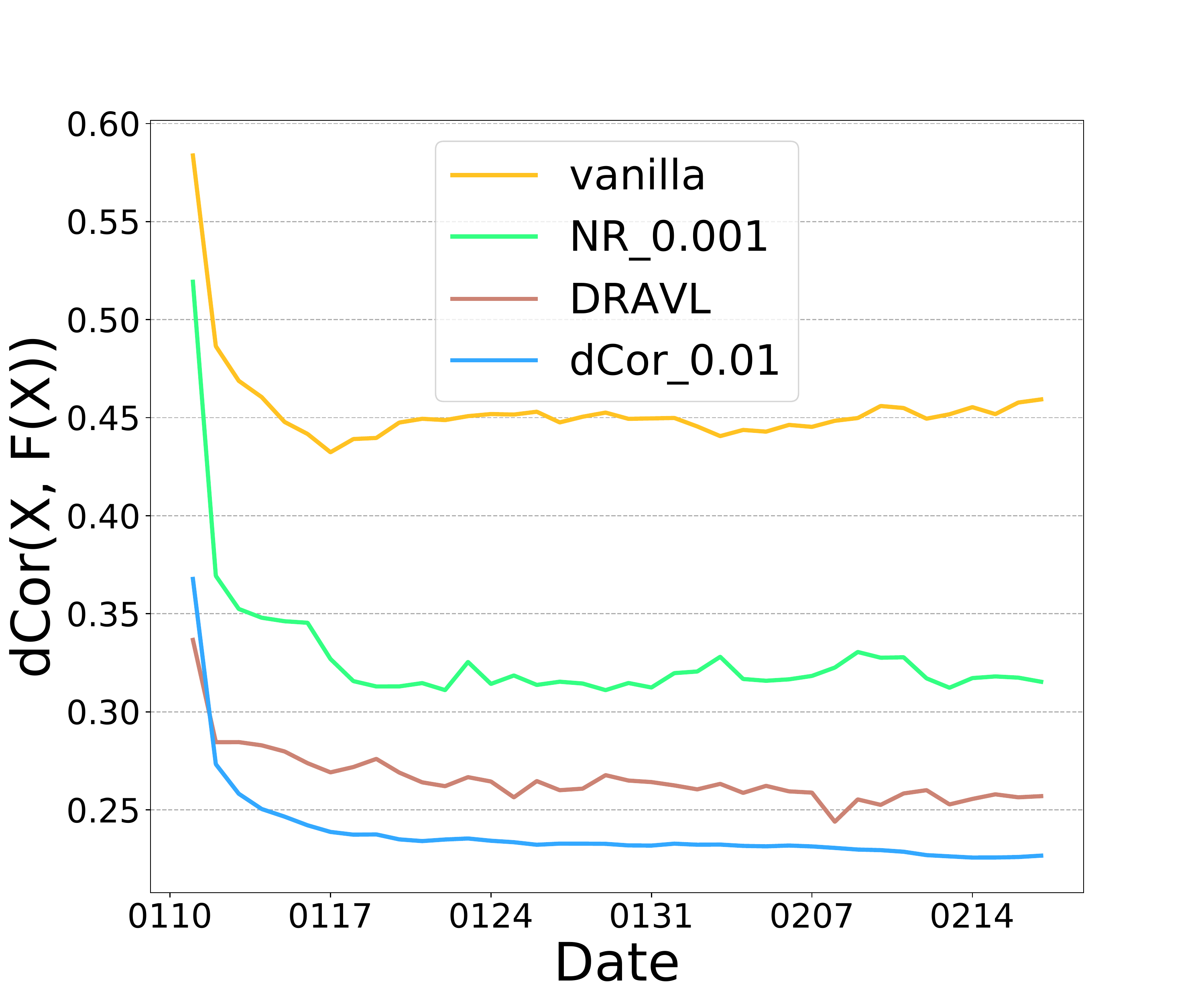}
    \caption*{(a)}
  \end{minipage}
  \begin{minipage}[b]{0.33\linewidth}
  \centering
    \includegraphics[width=\linewidth]{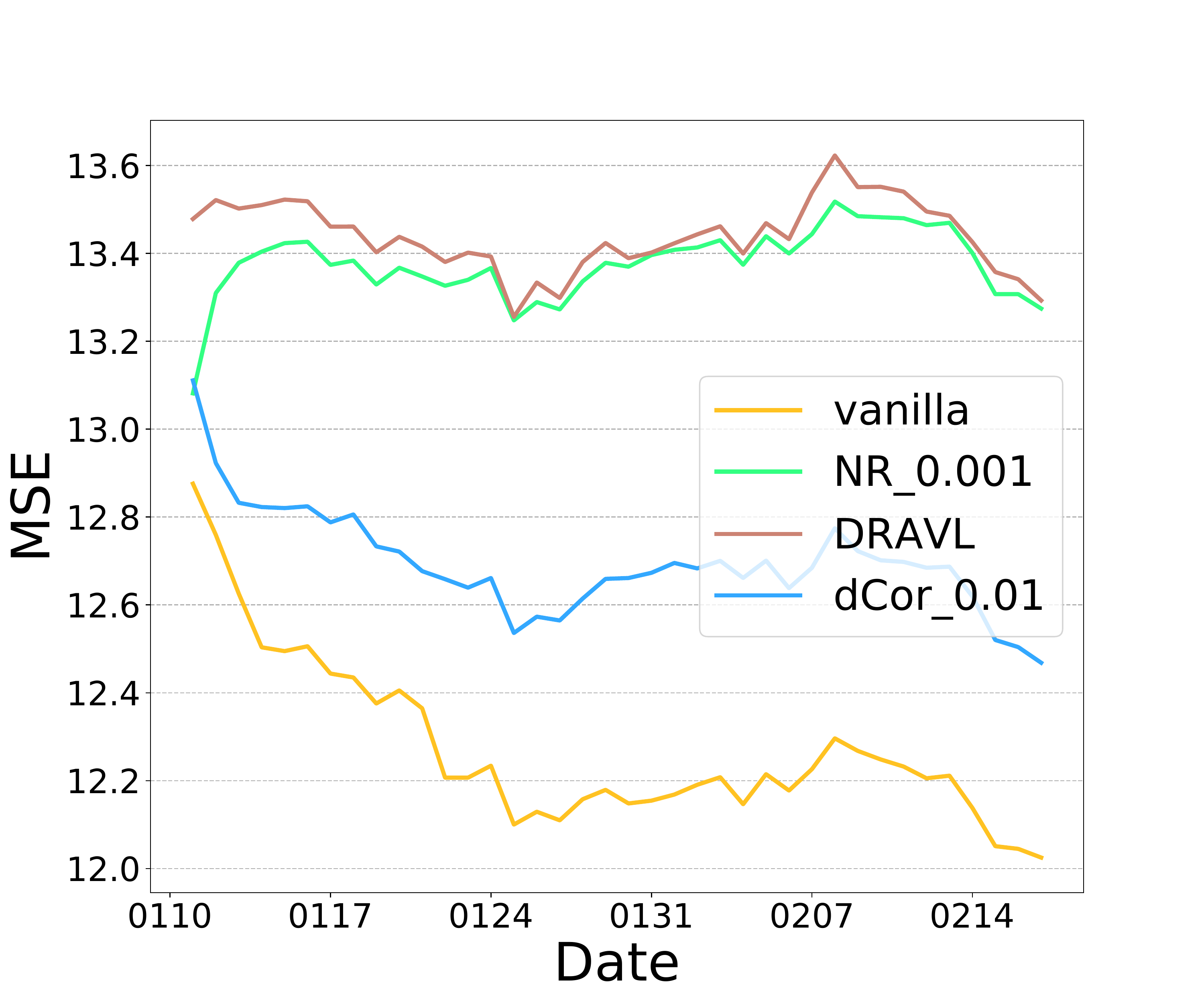}
    \caption*{(b)}
  \end{minipage}
  \begin{minipage}[b]{0.33\linewidth}
  \centering
    \includegraphics[width=\linewidth]{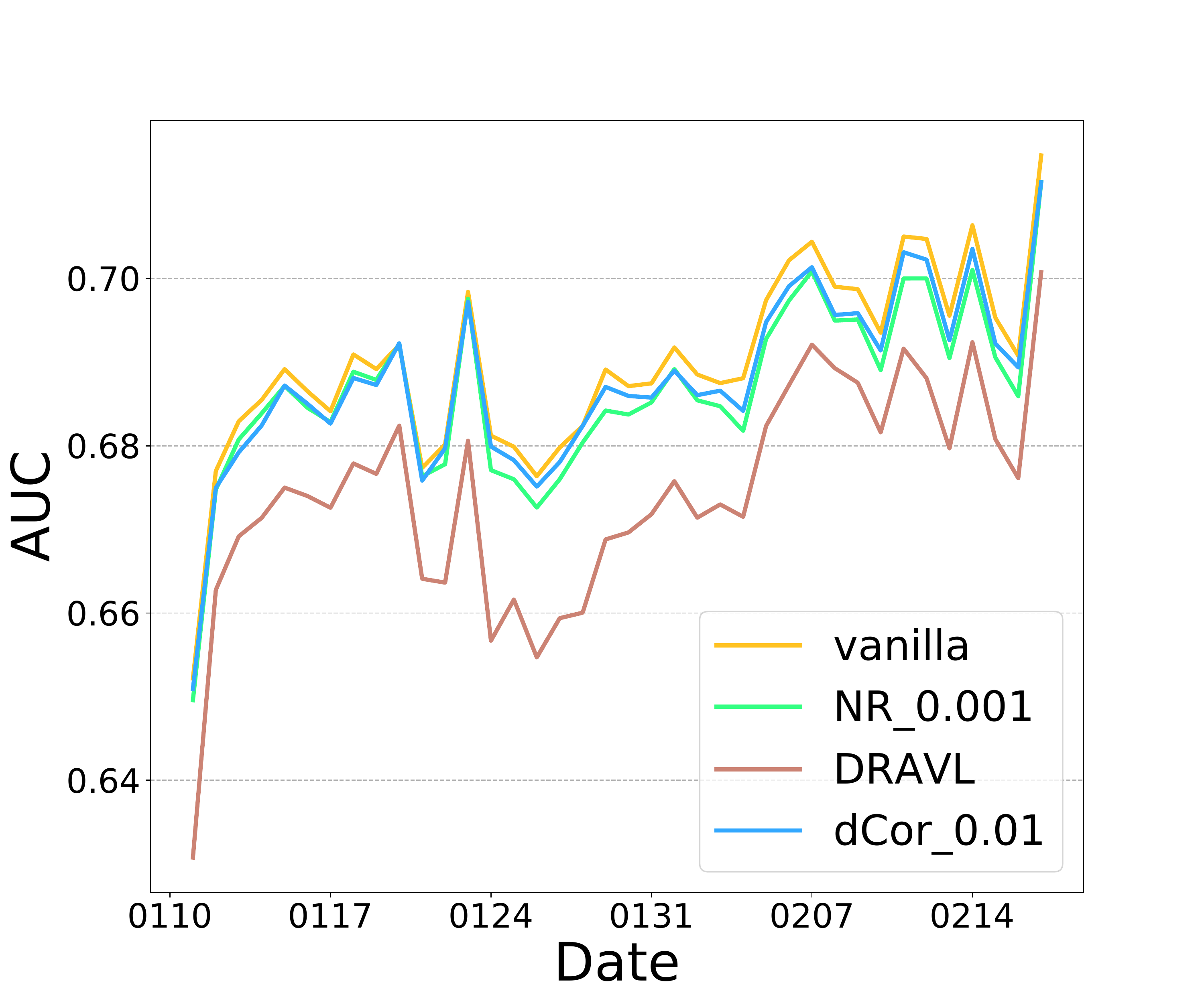}
    \caption*{(c)}
  \end{minipage}
  
    \begin{minipage}[b]{0.33\linewidth}
  \centering
    \includegraphics[width=\linewidth]{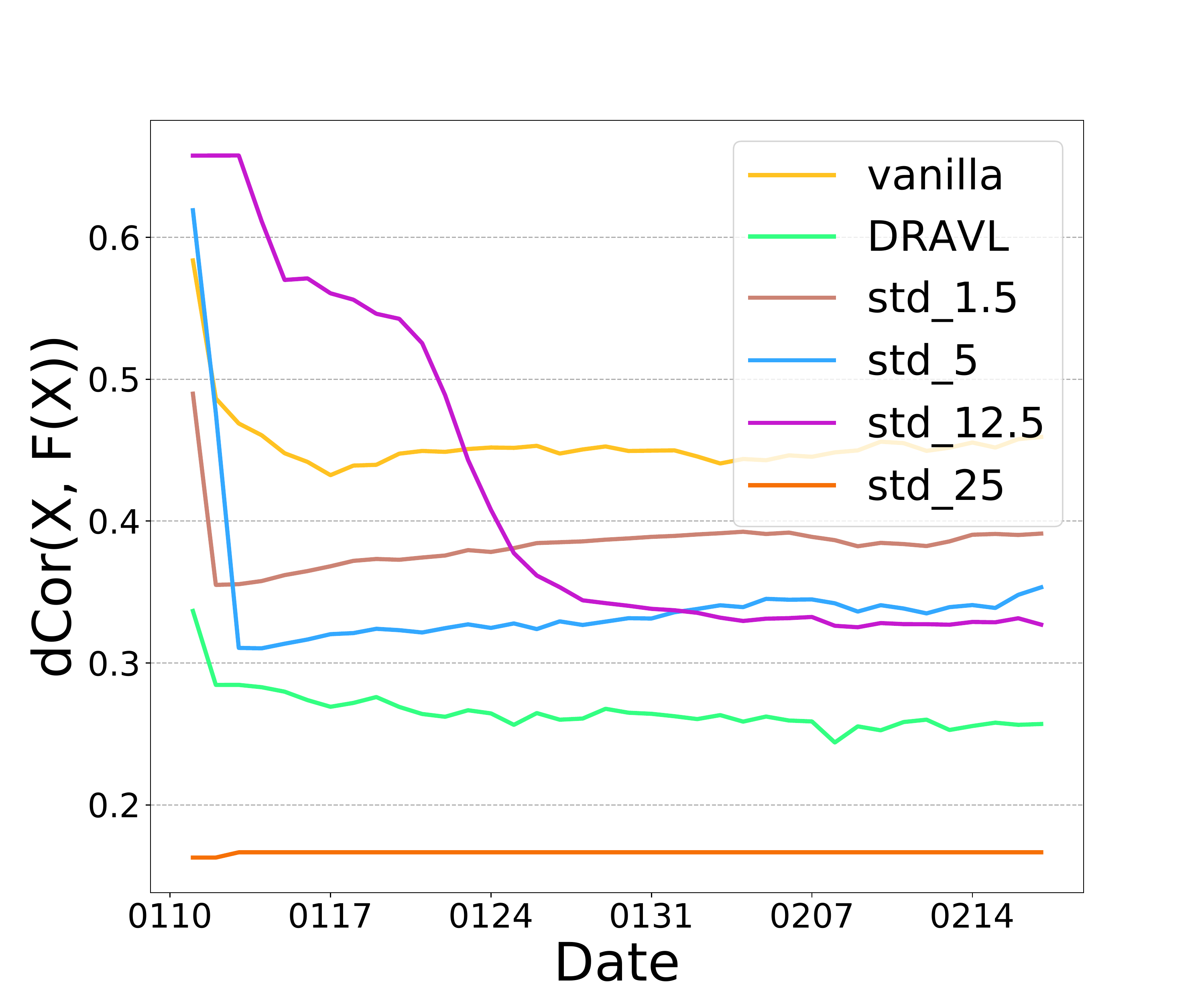}
    \caption*{(d)}
  \end{minipage}
  \begin{minipage}[b]{0.33\linewidth}
  \centering
    \includegraphics[width=\linewidth]{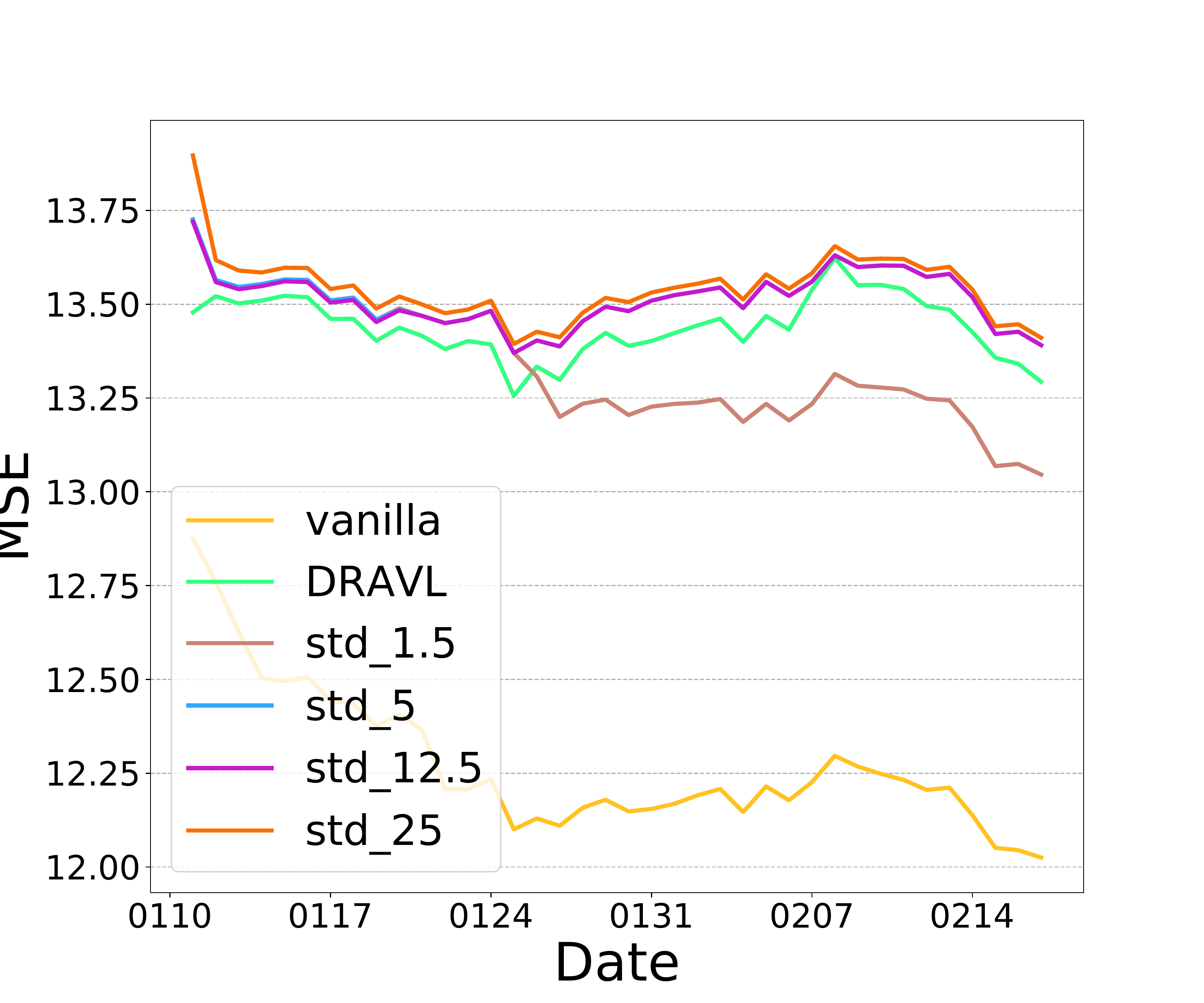}
    \caption*{(e)}
  \end{minipage}
  \begin{minipage}[b]{0.33\linewidth}
  \centering
    \includegraphics[width=\linewidth]{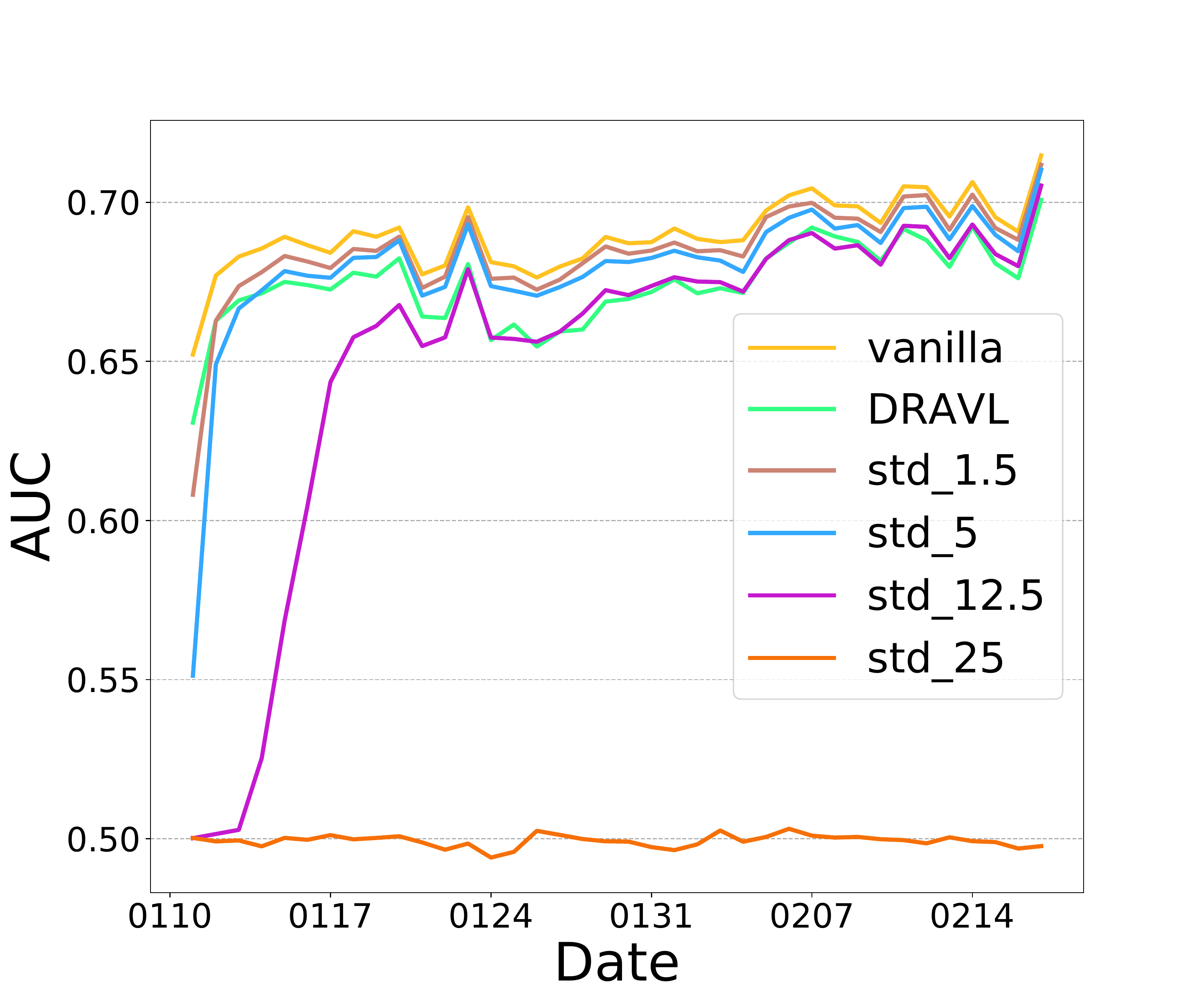}
    \caption*{(f)}
  \end{minipage}
  
 \caption{Figure (a), (b), and (c) demonstrate the effectiveness of DRAVL by comparing it with optimizing each module individually. The performance is evaluated by dCor($\mathcal{X}$, $\mathcal{F(X)}$) (Fig. (a)), MSE (Fig. (b)), and AUC (Fig. (c)). We compare the performance of the baseline (vanilla),  minimizing the dCor module only with $\alpha_d = 0.01$ (dCor\_0.01), NR module with $\alpha_n = 0.001$ (NR\_0.001), and our DRAVL with $\alpha_d = 0.01$ and $\alpha_r = 0.01$ 
 . Figure (d), (e), and (f) demonstrate the effectiveness of DRAVL by comparing it with adding noise to embedding. The performance is evaluated by dCor($\mathcal{X}$, $\mathcal{F(X)}$) (Fig. (d)), MSE (Fig. (e)), and AUC (Fig. (f)). We compare the performance of the baseline (vanilla),  DRAVL, 
 and adding $0$-mean Gaussian noise to embedding with different standard deviations.} 
 \label{fig:noise_reg_dcor} 
 \vspace{-0.175in}
\end{figure*}

We now demonstrate the effectiveness of optimizing all losses together, \ie{} DRAVL, by comparing it with individually optimizing each module. Figure~\ref{fig:noise_reg_dcor} (a) shows that using NR module alone can reduce more $dCor(\mathcal{X}, \mathcal{F(X)})$ than vanilla but its AUC is lower than dCor (Figure~\ref{fig:noise_reg_dcor} (c)). Figure~\ref{fig:noise_reg_dcor} (b) shows that dCor has lower MSE than NR, meaning NR preserves more privacy. This is unsurprising given NR is specifically designed to degrade the reconstruction quality. In addition, DRAVL helps gain the advantages of each module-it can reduce $dCor(\mathcal{X}, \mathcal{F(X)})$ and increase MSE simultaneously. Unfortunately, DRAVL also hurts AUC more than optimizing any of modules alone. However, in terms of finding the best overall tradeoff, DRAVL shows more promising results among all competitors: compared to vanilla, on average DRAVL increases MSE by $9.5\%$ and decreases the dCor by $41.44\%$ with a cost of AUC drop by only $2.25\%$.



We also compare DRAVL with a straightforward yet effective protection baseline: adding random noise to the cut layer embedding. We generate a random noise from a zero-mean Gaussian and add it to the embedding. We only tune the standard deviation of the Gaussian noise to control the noise strength. As shown in Figure~\ref{fig:noise_reg_dcor} (d), (e), and (f), with increasing the amount of noise added to the embedding, we can get a better privacy protection (lower dCor and higher MSE) but worse model utility (lower AUC). When noise strength is large enough (standard deviation $\geq 25$), the cut layer embedding is covered by the noise 
. As a result, the AUC drops to $0.5$, which is equivalent with a random guess. Overall, with a good control of the amount of the random noise added to the cut layer embedding, it might be an effective protection strategy.

We compare this noise perturbation with DRAVL in Figure~\ref{fig:noise_reg_dcor} (d), (e), and (f). DRAVL and noise perturbation method with $std=12.5$ can achieve similar AUC and MSE, since both models can make the reconstructed input be similar to the mean of the raw input. However, DRAVL reduces dCor more than the noise perturbation. 
Another drawback of perturbation based is that compared to DRAVL, empirically it is much harder to tune in order to find a good trade-off between model utility and privacy. 

\section{Related Work}
\label{sec:related-work}

\paragraph{Input Reconstruction Attack in FL.} Most of input reconstruction attacks are designed for Horizontal FL where a malicious server can reconstruct raw input from gradients sent from clients. Zhu et al. showed that an honest-but-curious server can jointly reconstruct raw data and its label from gradient on a 4-layer CNN~\cite{zhu2019dlg}. Geiping et al. extended the attack on deep models with ability to reconstruct high-resolution images~\cite{geiping2020inverting}. Yin et al. proposed the state-of-the-art method that is able to reconstruct high-resolution images in batches (in contrast to single-image optimization) from averaged gradient in a deep model~\cite{yin2021see}. We do not include those attacks in our experiments because they are designed for Horizontal FL and cannot be applied to Vertical FL. 
Luo et al. \cite{featureinference2021} studied the feature inference problem in the settings of vFL. The biggest differences with DRAVL is that they leverage prediction outputs in the prediction/inference stage of vFL to conduct the feature inference attacks. However, when some specific conditions are satisfied, e.g. the number of classes is large or the active party's features and the passive party's are highly correlated, their attack methods can infer the passive party's features well.   

\paragraph{Privacy-enhancement in FL.} 
There are mainly three categories of approaches to enhance privacy within existing FL framework: \textbf{1)} cryptography methods such as \textit{Secure Multi-party Computation} \cite{agrawal2019quotient, du2004privacy, bonawitz2017practical, nikolaenko2013privacy} and \textit{homomorphic encryption}\cite{aono2017privacy, sathya2018review}; \textbf{2)} system-based methods such as \textit{Trusted Execution Environments} \cite{subramanyan2017formal, tramer2018slalom}; \textbf{3)} perturbation methods such as randomly perturbing the communicated message \cite{abadi2016deep, mcmahan2017learning}, shuffling the messages \cite{erlingsson2019amplification, cheu2019distributed}, reducing message's data-precision, compressing and sparsifying the message \cite{zhu2019dlg}. 

\section{Conclusion}

In this paper, we design a defense framework that mitigates input leakage problems in Vertical FL. Our framework contains three modules: adversarial reconstruction, noise regularization, and distance correlation minimization. Those modules can not only be employed individually but also applied together since they are independent to each other. We conduct extensive experiments on a industrial-scale online advertising dataset to show that our framework is effective in protecting input privacy while maintain a reasonable model utility.  We urge the community to study more about privacy leakage problems in the context of Vertical FL, and to continue efforts to develop
more defenses against input reconstruction attacks and provide robustness against malicious parties.



\clearpage
\clearpage
\bibliography{references}
\bibliographystyle{icml2021}

\end{document}